\documentclass[letterpaper,10pt,conference]{IEEEtran}

\IEEEoverridecommandlockouts

\usepackage[
    letterpaper,
    left=0.75in,
    right=0.75in,
    top=0.85in,
    bottom=0.75in
]{geometry}

\usepackage[utf8]{inputenc}
\usepackage[T1]{fontenc}

\usepackage{textcomp}
\DeclareUnicodeCharacter{FFFD}{}
\usepackage{graphicx}
\usepackage{CJKutf8}
\usepackage{amsmath,amssymb}
\usepackage{CJKpunct}
\usepackage{booktabs,tabularx,multirow,array}
\usepackage{siunitx}
\usepackage[caption=false,font=footnotesize]{subfig}
\usepackage{xurl}
\usepackage{balance}

\title{\LARGE \bf
Reinforcement Learning for the Full Strawberry Harvesting Process: Obstacle Separation, Detachment, and Placement
}

\author{Changyou Miao$^1{^{,2}}$, Teng Li$^1$ and Ya Xiong$^1{^{,2}}$*
\thanks{This work was supported by the Haidian District Bureau of Agriculture and Rural Affairs, the Innovation Ability Project of Beijing Academy of Agricultural and Forestry Sciences (BAAFS) (KJCX20240321), the Outstanding Youth Foundation of BAAFS (YKPY2025007), and the NSFC Excellent Young Scientists Fund (overseas). (*\textit{Corresponding Author: Ya Xiong, \tt\small yaxiong@nercita.org.cn})}
\thanks{$^1$The Intelligent Equipment Research Center, Beijing Academy of Agriculture and Forestry Sciences, Beijing 100097, China.}%
\thanks{$^{2}$College of Engineering, Huazhong Agricultural University, Wuhan 430070, China}%
}

\begin{document}

\typeout{Paper size: \the\paperwidth\space x \the\paperheight}
\begin{CJK*}{UTF8}{gbsn}
\CJKpunctstyle{kaiming}
\relax

\maketitle
\thispagestyle{empty}
\pagestyle{empty}

\begin{abstract}
Severe occlusions and deformable plant structures introduce complex contact dynamics that challenge robotic strawberry harvesting. A policy-driven reinforcement learning (RL) framework with heuristic phase coordination was developed, in which obstacle separation, fruit detachment, and placement were formulated as a sequential decision-making task. A shared interaction-aware policy generated Cartesian motions across all task phases, while lightweight heuristic logic coordinated task progression and gripper events. A shared structured observation space was used to represent target, obstacle, end-effector, and task-context information. A hierarchical architecture combined the high-level policy with low-level Cartesian impedance control for compliant interaction. To support zero-shot sim-to-real transfer, feasibility-first observation alignment and domain randomization were adopted. The policy achieved success rates of 89.7\% in simulation and 82.0\% in real-world experiments. As the occlusion level increased from 1 to 5, the average execution time increased from 12.99 s to 21.73 s, reflecting greater interaction complexity. These results demonstrated effective transfer of interaction-aware harvesting behaviors to a structurally different robotic platform.
\end{abstract}

\section{Introduction}

Autonomous strawberry harvesting remains challenging because of the highly unstructured nature of agricultural environments. Unlike structured industrial settings, strawberry plants exhibit clustered fruit distributions, intertwined ripe and unripe berries, and dense, deformable foliage. Severe occlusion and complex plant geometries further increase perception and manipulation difficulty, often reducing harvesting success and increasing fruit damage in real-world applications. Most existing harvesting systems adopt a modular perception--planning--control pipeline, in which ripe fruits are first detected using vision algorithms, followed by predefined motion planning and rule-based grasping~\cite{R1,R2}. Although these approaches can be effective in relatively structured scenarios, their performance often deteriorates under severe occlusion and environmental disturbances. When target fruits are obstructed by leaves or unripe berries, passive collision avoidance or forced grasping motions are typically used, while the surrounding environment is rarely manipulated to improve harvesting feasibility~\cite{R3,R19}. Consequently, existing systems remain limited in their ability to actively clear obstacles before fruit detachment.

Reinforcement learning (RL) provides a promising alternative by formulating harvesting as a sequential decision-making problem. Through closed-loop state--action--reward interactions, long-horizon behaviors can be learned without requiring explicit models of complex plant interactions. However, the application of RL to agricultural harvesting remains constrained by deformable plant contact, severe occlusion, the limited feasibility of real-world training, and sim-to-real discrepancies~\cite{R6}.

\begin{figure}[t]
  \centering
  \includegraphics[width=0.95\columnwidth]{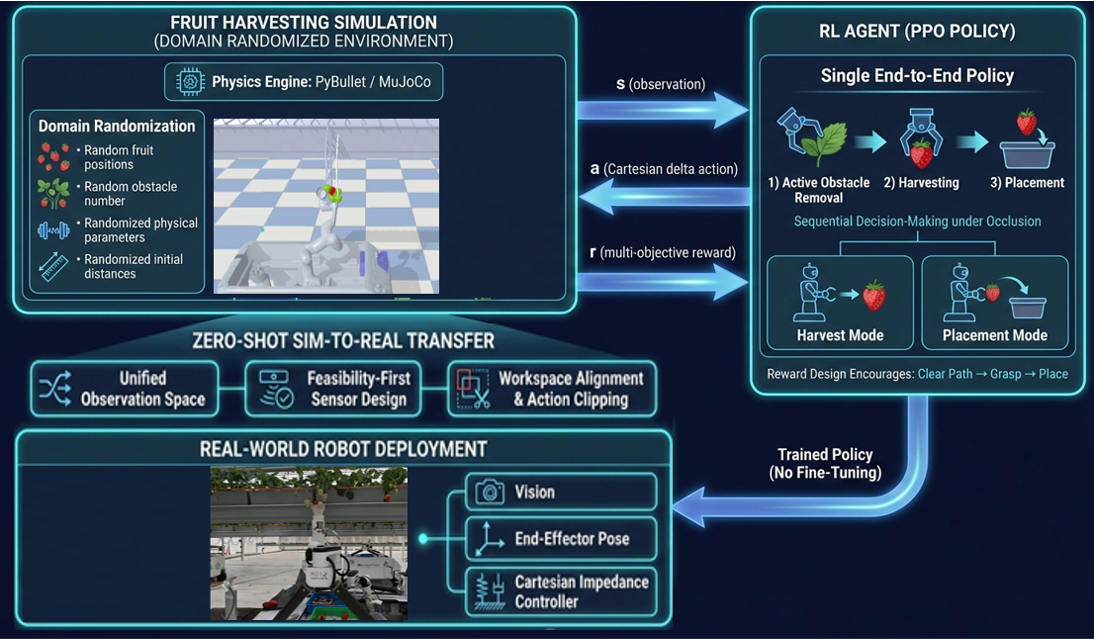}
  \caption{System overview of the sim-to-real framework. A PPO policy trained with domain randomization generates Cartesian increments for obstacle-aware harvesting and transfers zero-shot to the real robot.}
  \label{fig:fig1}
\end{figure}

To address these challenges, a policy-driven reinforcement learning framework with heuristic phase coordination was developed for occlusion-aware strawberry harvesting, as illustrated in Fig.~\ref{fig:fig1}. The obstacle-separation, fruit-detachment, and placement workflow was formulated as a sequential decision-making problem. A shared interaction-aware policy generated Cartesian motions across all task phases, while lightweight heuristic logic coordinated phase transitions and gripper events.

The main contributions are summarized as follows:
\begin{itemize}
\item \textbf{Policy-driven interaction-aware harvesting:} A shared interaction-aware policy was developed to generate obstacle-separation, fruit-detachment, and placement motions, while heuristic phase coordination organized task progression without replacing policy-based motion control.
\item \textbf{Feasibility-oriented sim-to-real alignment:} A shared structured observation space was used across task phases and aligned between simulation and real-world deployment to support zero-shot transfer.
\item \textbf{Real-world validation:} The proposed framework achieved an 82.0\% success rate in an elevated strawberry cultivation environment without real-world fine-tuning.
\end{itemize}

\section{Related Work}

\subsection{Reinforcement Learning in Agricultural Manipulation}

Reinforcement learning (RL) has recently been applied to agricultural manipulation. In strawberry harvesting, policies trained in \textit{FruitGym}, a MuJoCo-based environment incorporating large-scale domain randomization and Dormant Ratio Minimization, were deployed on a Franka Panda using Cartesian impedance control~\cite{R8}. Although effective sim-to-real transfer was demonstrated, the task primarily focused on target reaching and grasping rather than active manipulation of occluding plant elements~\cite{R23}. Beyond strawberry harvesting, a 7-DoF manipulator was trained for grape pruning using proximal policy optimization (PPO) by Yandún \textit{et al.}, with perception outputs incorporated into the state representation~\cite{R10}. Similarly, detected banana targets were included in the RL state by Lin \textit{et al.}, while the policy primarily learned inverse kinematics solutions~\cite{R11}. These studies demonstrated the feasibility of using RL as a motion generator conditioned on structured perception inputs. However, active obstacle separation and shared policy control across obstacle separation, fruit detachment, and placement under severe occlusion remain underexplored~\cite{R9}.

\subsection{Reinforcement Learning for Manipulation in Unstructured Environments}

Beyond agriculture, deep reinforcement learning has demonstrated strong potential for manipulation in cluttered or deformable environments. Deng \textit{et al.} applied RL to pushing and picking in cluttered tabletop scenes, where target objects were separated through physical interaction without explicit motion planning~\cite{R12}. Sim-to-real RL for deformable-object manipulation was investigated by Matas \textit{et al.}, and simulation-trained policies were transferred to real-world cloth manipulation tasks~\cite{R13}. The \textit{SoftGym} benchmark was introduced by Lin \textit{et al.} to study long-horizon control of deformable objects~\cite{R14}. However, these studies primarily addressed generic manipulation in controlled environments or tasks with single-stage objectives. Agricultural harvesting additionally involves severe occlusion, interleaved fruit distributions, and deformable plant structures, requiring coordinated interaction across multiple task phases. Shared policy control of sequential behaviors, including active obstacle separation, fruit detachment, and placement, remains insufficiently investigated~\cite{R24}.

\subsection{Position of This Work}

In contrast to prior agricultural RL methods that primarily focused on target reaching, grasping, or motion generation from structured perception inputs, this work formulated strawberry harvesting as an occlusion-aware sequential decision-making problem involving active environmental interaction. Occluding fruits and plant structures were manipulated to create accessible harvesting space before fruit detachment. An interaction-aware policy generated Cartesian motions across obstacle separation, fruit detachment, and placement, while heuristic phase coordination organized task progression and gripper events. Therefore, the contribution of this work lies in extending policy-based agricultural manipulation from isolated target-directed operations to active obstacle separation within a complete harvesting workflow, rather than in the standalone application of RL or Cartesian impedance control.

\section{Method}

The proposed framework integrated visual perception, policy-based Cartesian motion generation, heuristic phase coordination, and compliant robot execution for occlusion-aware strawberry harvesting. The perception and control architecture, simulation environment, observation and action spaces, and policy-learning method are described in the following subsections.

\subsection{Perception-Driven Policy Architecture}

A hierarchical control architecture was adopted to decouple policy-level Cartesian decision-making from robot-specific execution. Visual perception served as a supporting component that provided fruit-level semantic and geometric information for policy inference. Specifically, a YOLOv11-based SRR-Net estimated strawberry category, instance mask, spatial position, and maturity through parallel detection, segmentation, and ripeness-regression heads. RetinexNet-based illumination normalization was further applied to reduce lighting-induced color distortion under greenhouse conditions~\cite{R_LightStrawberry2026}.The detected fruit states were fused with robot proprioceptive information to form the structured observation vector described in the following subsection. Based on this input, the interaction-aware policy generated Cartesian motion increments, while heuristic logic coordinated phase transitions and gripper events. The resulting Cartesian commands were converted into joint commands through inverse kinematics and executed using Cartesian impedance control to support compliant interaction with deformable plant structures~\cite{R25}. This decoupling reduced the dependence of policy inference on platform-specific kinematics and supported transfer across structurally different robotic platforms.

\subsection{Simulation Environment and Domain Randomization}

A PyBullet-based simulation environment was constructed to model robot motion and contact interactions during strawberry harvesting. The simulated system consisted of a mobile base, a 7-DoF Franka Panda manipulator, a V-shaped elevated cultivation rack, and plant models containing ripe fruits, unripe fruits, and stems. Fruit bodies and stems were represented as separate links, while ripe and unripe fruits were assigned dark red and green colors, respectively, to approximate the primary visual cue used for maturity recognition. To approximate plant compliance, each stem was connected to the cultivation rack using a point-to-point spring--damper constraint, allowing oscillation and rebound during contact. Domain randomization was applied to plant geometry, occlusion complexity, and physical parameters~\cite{R20}.

\textbf{(1) Structural Randomization of Plant Geometry:}
Stem joint angles were independently sampled from a uniform distribution within $\pm 0.15$ rad (approximately $\pm 8.6^\circ$), corresponding to approximately 30\% of the original URDF joint limit of $\pm 0.523$ rad. This range provided plant-pose diversity while avoiding unrealistic deformation. The lateral displacement of the stem tip is approximated as

\begin{equation}
\Delta x \approx L \sin(0.15),
\end{equation}
where $L$ denotes the stem length. For $L=0.25~\mathrm{m}$, the resulting displacement was approximately $3.7~\mathrm{cm}$, which was sufficient to alter occlusion relationships and interaction paths.

\textbf{(2) Occlusion Complexity Randomization:}
During each episode, 1--5 unripe strawberries were randomly generated around the target fruit to produce different occlusion levels. This setting exposed the policy to varying interaction complexity and encouraged active obstacle-separation behavior during training.

\textbf{(3) Physical Parameter Perturbation:}
Spring--damper parameters were randomized at each episode reset to approximate variations in plant stiffness and contact dynamics. The maximum constraint force was sampled within 48--72 N around a nominal value of 60 N, while the error reduction parameter (ERP) was sampled within 0.18--0.22 around a nominal value of 0.2. Additional parameters, including mass, damping coefficients, contact friction, and initial fruit poses, were also perturbed within bounded ranges. These perturbations increased physical diversity while maintaining numerical stability.

\subsection{Observation Space}

A compact 32-dimensional shared structured observation space was designed to support obstacle-aware decision-making across obstacle separation, fruit detachment, and placement. Let the tool center point (TCP) and target-fruit positions be denoted by $\mathbf{p}_{\mathrm{tcp}}$ and $\mathbf{p}_{\mathrm{tgt}}$, respectively. Their Euclidean distance is defined as

\begin{equation}
d_t =
\left\lVert
\mathbf{p}_{\mathrm{tcp}}
-
\mathbf{p}_{\mathrm{tgt}}
\right\rVert_2,
\label{eq:distance}
\end{equation}

The observation vector consisted of four components:

\begin{itemize}
    \item \textbf{End-effector state (8D):}
    The TCP position $[x,y,z]$, orientation representation
    $[q_x,q_y,q_z,q_{\mathrm{pitch}}]$, and target-relative distance
    $d_{\mathrm{target}}$ in the world frame were included.

    \item \textbf{Target state (5D):}
    The target-relative position $\Delta\mathbf{p}_{\mathrm{target}}$,
    Euclidean distance $d_t$, and distance change $\Delta d_t$ were
    included.

    \item \textbf{Obstacle representation (15D):}
    The $K=3$ nearest obstacles were encoded, with each obstacle
    represented by a five-dimensional feature vector containing its
    relative position, binary ripeness encoding indicating ripe or
    unripe fruit, and distance to the TCP.

    \item \textbf{Task context (4D):}
    The relative drop-box position and a binary target-fruit
    grasp-status flag were included.
\end{itemize}

\subsection{Action Space}

Cartesian incremental control was adopted instead of joint-space control to reduce dependence on platform-specific kinematics and improve policy-learning stability. The four-dimensional normalized action is defined as

\begin{equation}
\mathbf{a}_t =
\left[
a_{x,t},
a_{y,t},
a_{z,t},
a_{\mathrm{pitch},t}
\right]
\in [-1,1]^4,
\label{eq:action_space}
\end{equation}
where the first three components represent translational actions and the fourth component represents pitch rotation. The normalized action was mapped to TCP motion increments as

\begin{equation}
\begin{aligned}
\Delta \mathbf{p}_{\mathrm{tcp},t}
&=
\alpha_{\mathrm{pos}}
\left[
a_{x,t},
a_{y,t},
a_{z,t}
\right]^{\mathrm{T}}, \\
\Delta \theta_{\mathrm{pitch},t}
&=
\alpha_{\mathrm{rot}}a_{\mathrm{pitch},t},
\end{aligned}
\label{eq:action_mapping}
\end{equation}
where $\alpha_{\mathrm{pos}}$ and $\alpha_{\mathrm{rot}}$ denote the translational and rotational scaling factors, respectively. This mapping bounded the motion increments and remained compatible with Cartesian impedance execution. Gripper closure and release were coordinated by the heuristic phase logic and were not included in the continuous action space.

\section{Policy Learning}

This section describes the task organization, curriculum strategy, and composite reward used to train the interaction-aware policy. Proximal policy optimization (PPO)~\cite{R21} was adopted for continuous Cartesian control. Unlike conventional modular pipelines~\cite{R15}, the same policy generated Cartesian motions throughout the harvesting workflow, while heuristic logic coordinated task transitions and gripper events.

\subsection{Policy Learning Framework}

Each harvesting episode consisted of two task modes: harvesting and placement. The harvesting mode included active obstacle separation and fruit detachment. When surrounding fruits obstructed the approach to the target, the policy generated interaction motions to progressively create accessible harvesting space before the final approach. The number, direction, and order of pushing motions emerged through policy optimization rather than being specified by predefined motion sequences. During execution, the spatial relationships among the end-effector, target fruit, and surrounding obstacles were continuously updated through the structured observation vector. The policy adapted its Cartesian actions to changes caused by plant contact and obstacle displacement. Heuristic phase coordination evaluated whether the target neighborhood had been sufficiently cleared and coordinated the grasp event, while the approach and interaction motions remained controlled by the learned policy.

After successful fruit detachment was confirmed, the task context was switched to the placement mode, in which the drop-box position served as the target reference. The policy then generated Cartesian motions based on the relative position between the end-effector and the drop box, while avoiding unnecessary contact with surrounding plant structures. The release event was triggered after the placement condition was satisfied. This organization preserved policy-based motion control while coordinating task transitions and gripper events.

\subsection{Curriculum Learning}

To accelerate convergence and stabilize policy training, a two-part curriculum strategy was applied.

\begin{itemize}
    \item \textbf{Distance curriculum:}
    During early training, the TCP was initialized 15\,cm from the target fruit. The initialization distance was then gradually increased to 1.0\,m as training progressed.

    \item \textbf{Mid-state initialization:}
    With a predefined probability, episodes were initialized from a post-grasp state in which the target fruit had already been grasped, thereby providing sufficient training samples for placement.
\end{itemize}

This strategy reduced the imbalance between harvesting and placement experiences and facilitated learning of the complete long-horizon workflow.

\subsection{Composite Reward Design for the Harvesting Process}

To facilitate learning of interaction-aware behaviors, a composite reward function was designed to guide target approach, active obstacle separation, fruit detachment, and placement. The reward design followed three principles: (i) providing dense guidance for efficient convergence during target approach, (ii) encouraging physically meaningful interactions that cleared occluding fruits before the final approach, and (iii) penalizing unsafe contacts, premature approach, and excessive interaction forces. Through this reward structure, the shared interaction-aware policy was encouraged to create accessible harvesting space before fruit detachment and subsequently complete placement.

The total reward is defined as

\begin{equation}
R_t = r_{\text{dist}} + r_{\text{prog}} + r_{\text{orient}} 
+ r_{\text{push}} + r_{\text{seq}} 
+ r_{\text{success}} + r_{\text{safe}}
\label{eq:reward_total}
\end{equation}

\subsubsection*{Notation}

Let $\mathbf{p}_{\text{tcp}}, \mathbf{p}_{\text{tgt}} \in \mathbb{R}^3$ denote the TCP and target positions. 
The Euclidean distance is defined as

\begin{equation}
d_t = \left\lVert \mathbf{p}_{\text{tcp}} - \mathbf{p}_{\text{tgt}} \right\rVert_2
\label{eq:distance_def}
\end{equation}

The distance variation is defined as

\begin{equation}
\Delta d_t = d_{t-1} - d_t
\label{eq:delta_distance}
\end{equation}

The obstacle set is denoted as $\mathcal{O} = \{o_1,\dots,o_K\}$.


\paragraph{Distance Reward}

\begin{equation}
r_{\text{dist}} = w_{d1} \left(1 - \tanh(\lambda d_t)\right)
\label{eq:r_dist}
\end{equation}

Through this term, dense global attraction toward the target is provided.


\paragraph{Progress Reward}

\begin{equation}
r_{\text{prog}} =
\begin{cases}
\lambda_{\text{app}} \Delta d_t, & \Delta d_t > 0 \\
\lambda_{\text{ret}} \Delta d_t, & \Delta d_t \le 0
\end{cases}
\label{eq:r_prog}
\end{equation}

Oscillatory retreat behavior was discouraged by the asymmetric coefficients ($\lambda_{\text{ret}} > \lambda_{\text{app}}$).


\paragraph{Orientation Constraint}

Pitch deviation is defined as

\begin{equation}
\theta_{\text{dist}} = \theta_{\text{pitch}} - \theta_{\text{opt}}
\label{eq:theta_dist}
\end{equation}

The orientation reward is defined as

\begin{equation}
r_{\text{orient}} =
\begin{cases}
1 - \dfrac{|\theta_{\text{dist}}|}{\delta_{\text{tol}}}, 
& |\theta_{\text{dist}}| \le \delta_{\text{tol}} \\
-\beta |\theta_{\text{dist}}|, 
& \text{otherwise}
\end{cases}
\label{eq:r_orient}
\end{equation}

Through this term, the policy was biased toward kinematically feasible grasp configurations.


\paragraph{Obstacle Interaction}

For each obstacle $o_i$, its distance to the target fruit was defined as

\begin{equation}
d_{\mathrm{obs\text{-}tgt},t}^{(i)}
=
\left\lVert
\mathbf{p}_{\mathrm{obs},t}^{(i)}
-
\mathbf{p}_{\mathrm{tgt},t}
\right\rVert_2,
\label{eq:d_obs}
\end{equation}
where $\mathbf{p}_{\mathrm{obs},t}^{(i)}$ and
$\mathbf{p}_{\mathrm{tgt},t}$ denote the obstacle and target positions,
respectively. The obstacle-separation reward was defined as

\begin{equation}
r_{\mathrm{push},t}
=
C_{\mathrm{push}}
\sum_{o_i \in \mathcal{O}_{\mathrm{obs}}}
\mathbb{I}
\left(
d_{\mathrm{obs\text{-}tgt},t}^{(i)}
>
d_{\mathrm{safe}}
\right)
\left(
1-b_{i,t-1}
\right),
\label{eq:r_push}
\end{equation}
where $C_{\mathrm{push}}>0$ denotes the fixed separation reward, and
$b_{i,t-1}\in\{0,1\}$ indicates whether obstacle $o_i$ had previously
satisfied the separation condition. Once the reward was triggered,
$b_{i,t}$ was set to 1. Thus, each obstacle was rewarded only once, and
additional displacement beyond $d_{\mathrm{safe}}$ yielded no further
reward.

\paragraph{Sequential Constraint}

The target-neighborhood clearance condition was defined as

\begin{equation}
S_{\mathrm{clear},t}
\iff
\forall o_i \in \mathcal{O}_{\mathrm{obs}},
\quad
d_{\mathrm{obs\text{-}tgt},t}^{(i)}
>
d_{\mathrm{safe}}.
\label{eq:s_clear}
\end{equation}

The sequential penalty was defined as

\begin{equation}
r_{\mathrm{seq},t}
=
-
C_{\mathrm{seq}}
\mathbb{I}
\left(
\neg S_{\mathrm{clear},t}
\land
d_t < d_{\mathrm{near}}
\right),
\label{eq:r_seq}
\end{equation}
where $C_{\mathrm{seq}}>0$ denotes the penalty coefficient, $d_t$ denotes
the end-effector-to-target distance, and $d_{\mathrm{near}}$ denotes the
near-target threshold. This term penalized premature approach and
encouraged obstacle separation before fruit detachment.

\paragraph{Success Reward}

\begin{equation}
r_{\text{success}} =
\mathbb{I}_{\text{picked}}
\left(
R_{\text{base}} + \max(0, T_{\text{max}} - \eta t)
\right)
\label{eq:r_success}
\end{equation}

\paragraph{Safety Penalty}

\begin{equation}
r_{\text{safe}} =
-w_f \max(0, F_{\text{contact}} - F_{\text{limit}})
- w_c \mathbb{I}_{\text{collision}}
\label{eq:r_safe}
\end{equation}

Through this penalty, excessive contact force and unintended collisions were discouraged.

\section{Policy Training}

This section describes the simulation-based training setup and implementation details.

\subsection{Simulation Infrastructure}

Training was conducted using 16 parallel instances of the PyBullet environment described above on an NVIDIA RTX 6000 GPU. Parallel rollout collection increased training throughput and provided diverse interaction trajectories for policy optimization.

\subsection{Algorithm and Network Architecture}

The interaction-aware policy was trained using PPO. PPO was selected for its empirical stability in continuous control tasks and compatibility with parallel simulation. The implementation was based on Stable-Baselines3. An actor--critic architecture was adopted, in which a multilayer perceptron (MLP) with hidden-layer sizes of 512--512--256--512 was used as the feature extractor. The policy and value heads each consisted of two hidden layers with 256 units per layer.

\begin{table}[!t]
\centering
\scriptsize
\setlength{\tabcolsep}{3pt}
\caption{
Training configuration and sim-to-real performance. Domain randomization was applied to plant geometry ($\pm 0.15$ rad), contact dynamics (60\,N $\pm$ 20\%, ERP 0.2 $\pm$ 10\%), and occlusion levels (1--5 objects).
}
\label{tab_training}
\begin{tabularx}{\columnwidth}{l l >{\raggedright\arraybackslash}X}
\toprule
\textbf{Category} & \textbf{Parameter} & \textbf{Value} \\
\midrule

\multirow{2}{*}{Observation} 
& Obs dim & 32 \\
& Action dim & 4 ($\Delta x$, $\Delta y$, $\Delta z$, $\Delta pitch$) \\
\midrule

\multirow{2}{*}{Network} 
& Feature extractor & MLP [512,512,256,512] \\
& Policy / value heads & MLP [256,256] \\
\midrule

\multirow{5}{*}{PPO} 
& Learning rate & $1\times10^{-4}\rightarrow 5\times10^{-5}$ \\
& Batch / steps / epochs & 256 / 1024 / 8 \\
& $\gamma$ / $\lambda$ & 0.99 / 0.95 \\
& Clip / entropy & $0.2\rightarrow0.1$ / $0.05\rightarrow0.04$ \\
& Normalization & Observation/Reward enabled \\
\midrule

\multirow{4}{*}{Domain randomization} 
& Stem joint perturbation & $\pm 0.15$ rad ($\approx$ 30\%) \\
& Max constraint force & 60\,N $\pm$ 20\% \\
& ERP & 0.2 $\pm$ 10\% \\
& Occlusion range & 1--5 objects \\
\midrule

\multirow{3}{*}{Training scale}
& Parallel envs & 16 \\
& Total steps & 6.03M \\
& Training time & 7.5 hours \\
\midrule

\multirow{2}{*}{Deployment}
& Simulation success & 89.7\% \\
& Real-world success & 82\% \\
\bottomrule
\end{tabularx}
\end{table}

\begin{figure}[!htbp]
  \centering
\includegraphics[width=0.8\columnwidth]{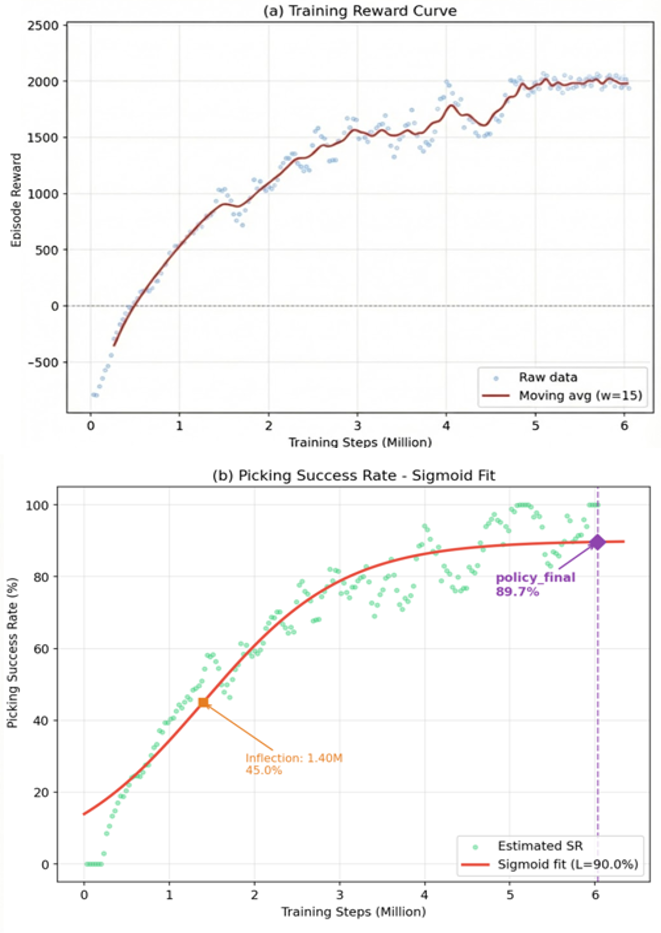}
  \caption{Convergence performance of the trained PPO policy in the strawberry harvesting task. 
(a) Episode reward over training steps, showing convergence within 6M steps and stabilization after 5.08M steps. 
(b) Picking success rate with sigmoid fitting, reaching approximately 89.7\%.}
  \label{fig:fig2}
\end{figure}

The learning rate was linearly decreased from $1\times10^{-4}$ to $5\times10^{-5}$, while the entropy coefficient was decreased from 0.05 to 0.04 to balance exploration and policy convergence. The PPO clipping range was linearly decreased from 0.2 to 0.1. The discount factor and generalized advantage estimation (GAE) parameter were set to $\gamma=0.99$ and $\lambda=0.95$, respectively. The main training configuration is summarized in Table~\ref{tab_training}.

Training converged after approximately 6.03 million timesteps. As shown in Fig.~\ref{fig:fig2}, the average episode reward stabilized around 2357 during the late training stage, while the simulated harvesting success rate reached 89.7\%. The value loss decreased from 0.139 to 0.027, corresponding to an approximately 81\% reduction, while the policy loss and entropy loss remained near $-0.006$ and $-4.60$, respectively. These training diagnostics indicated stable policy optimization. The trained policy parameters were subsequently fixed for real-world deployment.

\subsection{Ablation Study}

Two ablation experiments were conducted to evaluate the effects of the obstacle-separation mechanism and curriculum learning. For a fair comparison, the same PPO network architecture, four-dimensional action space, placement procedure, evaluation conditions, and maximum episode length of 900 steps were used across the corresponding experimental groups. The specific components removed in each ablation are described below.

\subsubsection{Effect of Obstacle Separation}

The first ablation evaluated the combined contribution of heuristic obstacle-separation coordination and its associated reward guidance. In the baseline, separation activation, separation confirmation, the post-separation fruit-detachment gate, and the obstacle-separation-related reward terms were disabled. The remaining reward components, policy-based motion control, and placement procedure were kept unchanged. Thus, the shared interaction-aware policy remained responsible for Cartesian motion generation in both groups.

Five training repetitions were conducted using seeds 0--4, with each seed trained for 100,000 steps. The harvesting success rates with and without the obstacle-separation mechanism were 88\%/60\%, 85\%/59\%, 88\%/52\%, 93\%/58\%, and 86\%/56\%, respectively. The mean success rate increased from 57.0\% to 88.0\%, corresponding to an improvement of 31.0 percentage points. These results demonstrated that the obstacle-separation mechanism substantially improved harvesting reliability when obstacles were located near the target fruit.

\subsubsection{Effect of Curriculum Learning}

The second ablation evaluated early-stage training efficiency. In the uniform-training baseline, the distance curriculum and mid-state initialization were removed, while the complete obstacle-separation mechanism, reward design, placement control, and evaluation conditions remained unchanged. Five paired seeds were trained for 300,000 steps from random initialization, with checkpoints saved every 50,000 steps.

The convergence criterion was defined as the first training step at which the full-task success rate remained at or above 50\% for two consecutive checkpoints. The curriculum group satisfied this criterion for all five seeds, whereas the uniform-training group satisfied it for only two seeds. The normalized area under the success-rate curve (AUC) over 0--300,000 steps increased from 43.77\% to 64.33\%. These results indicated that curriculum learning improved early convergence and learning efficiency.

Overall, the obstacle-separation mechanism primarily improved final harvesting reliability, whereas curriculum learning mainly accelerated early-stage policy training.

\section{Sim-to-Real Transfer}

Zero-shot transfer was supported by preserving consistency in observation representation, action semantics, coordinate definitions, and workspace feasibility between simulation and the real system. The transfer strategy focused on task-specific consistency adaptations rather than introducing new normalization or low-level control techniques.

\begin{table*}[t]
\centering
\small
\setlength{\tabcolsep}{6pt}
\caption{Real-world execution time analysis under different occlusion levels.}
\label{tab_time_analysis}
\begin{tabular}{
c
S
S
S
S
S[table-format=1.2]
S[table-format=3.1]
S[table-format=2.2]
}
\toprule
\textbf{Obs.} 
& {\textbf{Percep. (s)}} 
& {\textbf{Sep. (s)}} 
& {\textbf{Approach (s)}} 
& {\textbf{Place. (s)}} 
& {\textbf{Infer. (ms)}} 
& {\textbf{Steps}} 
& {\textbf{Total (s)}} \\
\midrule
1 & 0.22 & 4.93 & 2.20 & 5.64 & 0.74 & 63.3  & \bfseries \num{12.99} \\
2 & 0.22 & 5.29 & 2.33 & 5.73 & 0.75 & 67.6  & \bfseries \num{13.57} \\
3 & 0.15 & 7.76 & 2.23 & 5.81 & 0.75 & 109.0 & \bfseries \num{15.95} \\
4 & 0.15 & 10.13& 3.67 & 6.04 & 0.74 & 117.0 & \bfseries \num{19.99} \\
5 & 0.08 & 9.56 & 5.77 & 6.32 & 0.76 & 143.0 & \bfseries \num{21.73} \\
\midrule
\textbf{Mean} 
& 0.16 
& 7.53 
& 3.24 
& 5.91 
& 0.75 
& 100.0 
& \bfseries \num{16.85} \\
\bottomrule
\end{tabular}
\end{table*}

\begin{figure*}[t]
  \centering
  \subfloat[Successful Grasping under a One-Obstacle Occlusion Scenario]{%
    \includegraphics[width=0.93\textwidth]{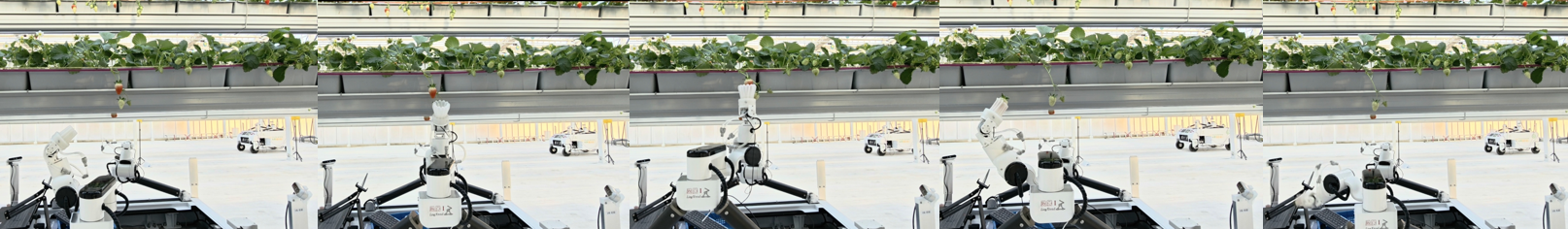}%
  }\par\medskip

  \subfloat[Successful Grasping under a Five-Obstacle Occlusion Scenario]{%
    \includegraphics[width=0.93\textwidth]{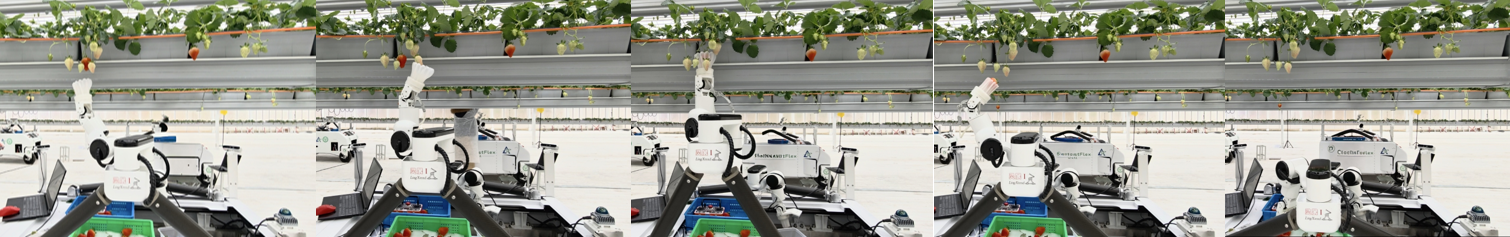}%
  }\par\medskip

  \subfloat[Failure Case: Slippage of the Target Strawberry during Grasping]{%
    \includegraphics[width=0.93\textwidth]{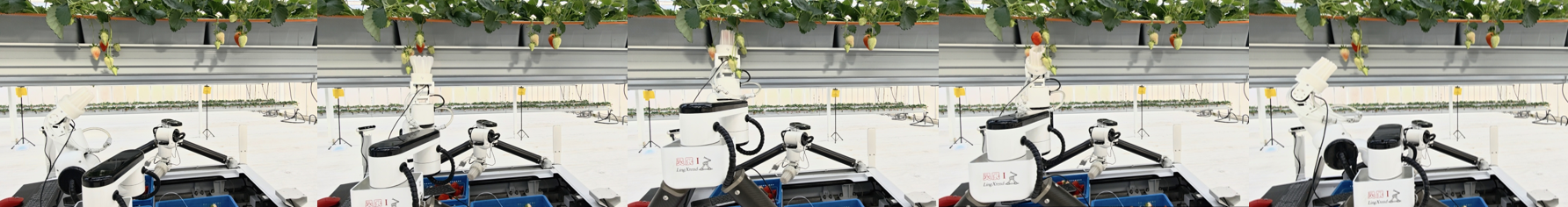}%
  }

  \caption{Key Frames of the Harvesting Test for Successful and Failed Cases}
  \label{fig:failures}
\end{figure*}

\subsection{Observation Alignment and Sim-to-Real Consistency}

A key challenge in sim-to-real deployment arose from inconsistencies between simulated and real observation representations. To address this issue, a \textit{Feasibility-First} principle was adopted. The shared structured observation space defined above was restricted to states physically measurable by the real system, while the simulation environment was adapted to reproduce the same observation semantics. Formally, let $\mathcal{S}^{\mathrm{real}}$ denote the set of physically measurable states. The aligned observation space $\mathcal{O}$ is defined as

\begin{equation}
\mathcal{O}
=
\arg\max_{O \subseteq \mathcal{S}^{\mathrm{real}}}
\mathcal{I}(O;\pi^*),
\label{eq:observation_alignment}
\end{equation}
where $\mathcal{I}(O;\pi^*)$ denotes the mutual information between a candidate observation subset $O$ and the optimal policy $\pi^*$. Under real-world sensing constraints, this formulation retained decision-relevant information while reducing dependence on variables unavailable during deployment.

Three alignment mechanisms were implemented to maintain consistency between simulation and real-world deployment.

\subsubsection{Geometric and Coordinate Consistency}

The policy was conditioned on the TCP pose expressed relative to the robot base. Differences in coordinate semantics between simulation and hardware could introduce out-of-distribution observations and produce invalid control commands. Therefore, the base reference frames, coordinate origins, and TCP definitions were aligned between the simulated and real systems. Kinematic consistency was verified using multiple canonical poses, for which the positional deviation between simulation and hardware remained within $5~\mathrm{mm}$.

\subsubsection{Action Semantics Consistency}

The normalized actions generated by the policy were transformed into executable TCP increments through a multi-stage processing pipeline comprising base scaling, distance-adaptive scaling, action smoothing, and directional correction. The same processing sequence and parameter definitions were reproduced during real-world deployment to preserve the action semantics learned in simulation. When a processed command exceeded the predefined safety limit, proportional safety scaling was applied to reduce its magnitude while preserving its direction.

\subsubsection{Observation Distribution Alignment}

During training, observations were normalized using the \texttt{VecNormalize} module in Stable-Baselines3, which maintained running estimates of the per-dimension means $\mu_i$ and variances $\sigma_i^2$. The stored training-time normalization statistics were reused during deployment to preserve consistency in observation preprocessing. All observation components were also kept consistent in physical meaning, dimensionality, computation method, and coordinate reference, thereby reducing semantic and statistical drift between simulation and real-world deployment.

\subsection{Workspace Constraints}

The executable workspace was defined as the intersection of the simulated Franka Panda workspace and the workspace of the LingXtend manipulator used in the real-world experiments. Cartesian commands outside this region were clipped to the corresponding workspace bounds, ensuring that all executed motions remained feasible for both robotic platforms and within predefined mechanical safety limits.

\subsection{Real-World Experimental Setup and Results}

\begin{figure}[!htbp]
  \centering
\includegraphics[width=1\columnwidth]{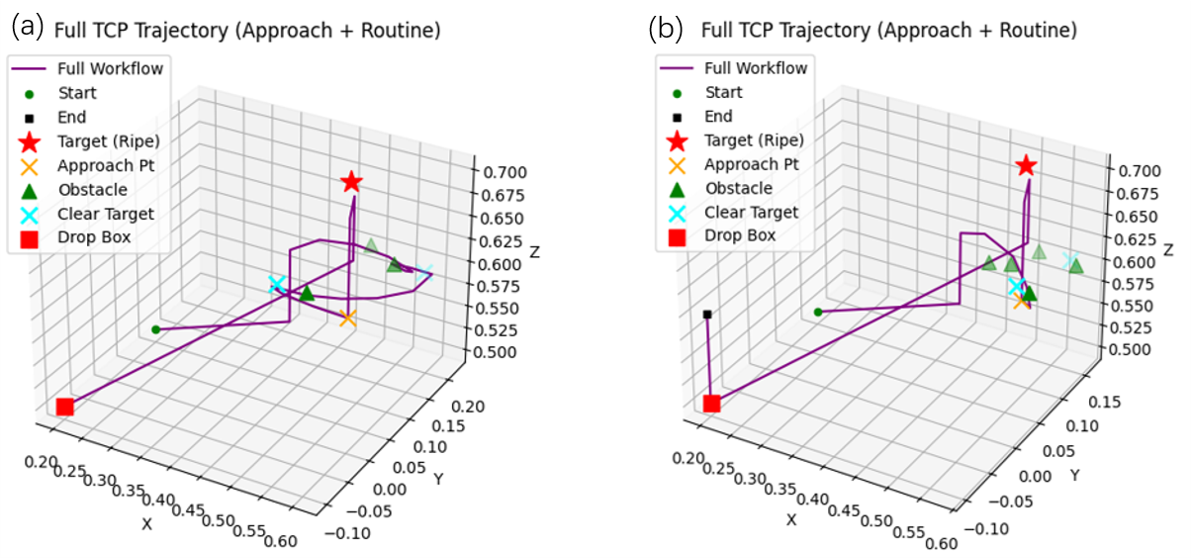}
  \caption{Harvesting trajectories under different occlusion levels.
(a) Harvesting test with three obstacles;
(b) Harvesting test with five obstacles.}
  \label{fig:fig4}
\end{figure}

\begin{figure}[!htbp]
  \centering
\includegraphics[width=0.95\columnwidth]{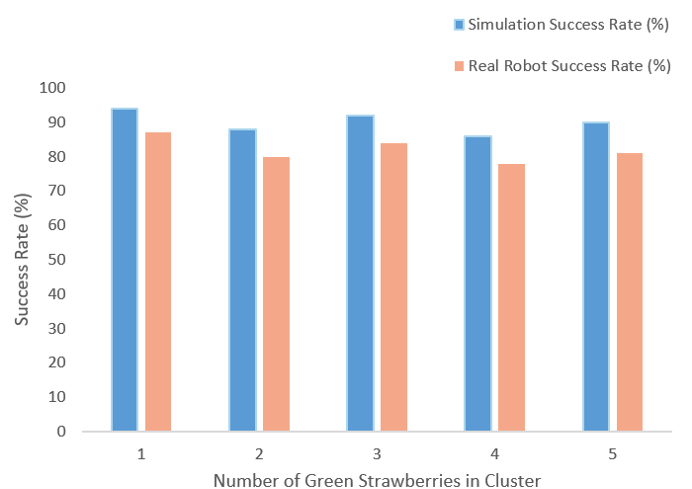}
  \caption{Comparison of harvesting success rates in simulation and real-world experiments across occlusion levels (1–5 obstacles). Each condition included 20 trials. Results showed consistent performance trends with a moderate sim-to-real gap.}
  \label{fig:fig5}
\end{figure}

Real-world experiments were conducted using the LingXtend hybrid serial--parallel manipulator~\cite{R_HybridArm_IROS2024}. A Cartesian impedance controller was implemented in ROS to support compliant interaction with deformable plant structures. Visual observations were acquired using an Intel RealSense D435 camera mounted near the robot base. The experiments were performed in an elevated strawberry cultivation system at a commercial greenhouse operated by Cuihu Agriculture Technology Co., Ltd., Beijing, China, where ripe and unripe fruits were distributed in an interleaved manner.

Five occlusion levels with 1--5 surrounding obstacles were evaluated. Twenty harvesting trials were conducted at each level, resulting in a total of 100 real-world trials. A trial was regarded as successful only when the complete harvesting sequence was accomplished: the occluding fruits were cleared from the approach path, the target strawberry was detached, and the fruit was transported to and released at a position $5~\mathrm{cm}$ above the drop box. Failure at any stage, including obstacle separation, fruit detachment, target tracking, or placement, resulted in an unsuccessful trial. The proposed framework completed the full harvesting sequence in 82 trials, yielding an overall success rate of 82.0\%. As summarized in Table~\ref{tab_time_analysis}, the average total execution time increased from 12.99~s to 21.73~s as the occlusion level increased from 1 to 5.

Among the 18 failed trials, target strawberries positioned close to the cultivation rack restricted feasible approach angles and reduced the available manipulation space. Insufficient stem bending also prevented fruit detachment despite appropriate end-effector positioning, as illustrated in Fig.~\ref{fig:failures}. In addition, displaced unripe strawberries occasionally rebounded because of plant elasticity, reintroducing occlusion and causing target loss or suboptimal approach motions. Harvesting trajectories under different occlusion levels are shown in Fig.~\ref{fig:fig4}, while the simulation and real-world success rates are compared in Fig.~\ref{fig:fig5}. Despite these failures, the results demonstrated the feasibility of zero-shot sim-to-real transfer without additional real-world fine-tuning.

\section{Discussion}

A direct quantitative comparison with existing strawberry-harvesting methods was not conducted because most previous studies evaluated isolated operations, such as target approach, grasping, or fruit detachment, whereas this work evaluated the complete workflow of obstacle separation, fruit detachment, and placement. Moreover, the spring--damper model did not fully reproduce the nonlinear deformation and contact dynamics of real strawberry plants, and experiments were conducted using only one robotic platform in a single elevated cultivation environment. Variations across cultivars, seasons, and cultivation systems, together with fruit and plant damage, were not systematically evaluated. Future work will include standardized full-process comparisons, broader field validation, more realistic plant models, and force- and damage-aware evaluation.

\section{Conclusion}

A policy-driven reinforcement learning framework with heuristic phase coordination was developed for occlusion-aware strawberry harvesting. The interaction-aware policy generated Cartesian motions throughout the harvesting workflow, while heuristic logic coordinated task transitions and gripper events. A hierarchical control architecture and consistency-oriented sim-to-real adaptations supported compliant execution and zero-shot deployment on a structurally different robotic platform. The policy achieved success rates of 89.7\% in simulation and 82.0\% in real-world experiments. As the occlusion level increased from 1 to 5, the average execution time increased from 12.99~s to 21.73~s, primarily because of longer obstacle-separation and target-approach stages, while perception and policy-inference times remained relatively stable. These results demonstrated the feasibility of transferring interaction-aware harvesting behaviors learned in simulation to real-world strawberry harvesting without additional policy fine-tuning.

\section*{Acknowledgment}
Figure~1 was created using a generative AI tool (Gemini) based on a text prompt. 
A demonstration video is available at:
\url{https://youtu.be/3X6CT0IuVKc}.

\balance
\bibliographystyle{IEEEtran}
\bibliography{references}

@article{R1,
author = {Xiong, Ya and Ge, Yuanyue and Grimstad, Lars and From, Pål J.},
title = {An autonomous strawberry-harvesting robot: Design, development, integration, and field evaluation},
journal = {Journal of Field Robotics},
volume = {37},
number = {2},
pages = {202-224},
doi = {https://doi.org/10.1002/rob.21889},
url = {https://onlinelibrary.wiley.com/doi/abs/10.1002/rob.21889},
eprint = {https://onlinelibrary.wiley.com/doi/pdf/10.1002/rob.21889},
year = {2020}
}

@article{R2,
  author  = {B. Arad and J. Balendonck and R. Barth and O. Ben-Shahar and Y. Edan and T. Hellstr{\"o}m and J. Hemming and P. Kurtser and O. Ringdahl and T. Tielen},
  title   = {Development of a Sweet Pepper Harvesting Robot},
  journal = {Journal of Field Robotics},
  volume  = {37},
  number  = {6},
  pages   = {1027--1039},
  year    = {2020}
}

@article{R3,
  author  = {J. Jun and J. Kim and J. Seol and H. I. Son},
  title   = {Towards an Efficient Tomato Harvesting Robot: 3D Perception, Manipulation, and End-Effector},
  journal = {IEEE Access},
  volume  = {9},
  pages   = {17631--17640},
  year    = {2021}
}

@inproceedings{R6,
  author    = {J. Tobin and R. Fong and A. Ray and J. Schneider and W. Zaremba and P. Abbeel},
  title     = {Domain Randomization for Transferring Deep Neural Networks from Simulation to the Real World},
  booktitle = {Proc. IEEE/RSJ Int. Conf. Intelligent Robots and Systems (IROS)},
  pages     = {23--30},
  year      = {2017}
}

@article{R8,
author = {E. Williams and A. Polydoros},
title = {Zero-Shot Sim-to-Real Reinforcement Learning for Fruit Harvesting},
journal = {arXiv preprint arXiv:2505.08458},
year = {2025}
}

@inproceedings{R9,
  author    = {E. Todorov and T. Erez and Y. Tassa},
  title     = {MuJoCo: A Physics Engine for Model-Based Control},
  booktitle = {Proc. IEEE/RSJ Int. Conf. Intelligent Robots and Systems (IROS)},
  pages     = {5026--5033},
  year      = {2012}
}

@inproceedings{R10,
  author    = {F. Yandun and T. Parhar and A. Silwal and D. Clifford and Z. Yuan and G. Levine and S. Yaroshenko and G. Kantor},
  title     = {Reaching Pruning Locations in a Vine Using a Deep Reinforcement Learning Policy},
  booktitle = {Proc. IEEE Int. Conf. Robotics and Automation (ICRA)},
  pages     = {2400--2406},
  year      = {2021}
}

@article{R11,
  author  = {G. Lin and P. Huang and M. Wang and Y. Xu and R. Zhang and L. Zhu},
  title   = {An Inverse Kinematics Solution for a Series-Parallel Hybrid Banana-Harvesting Robot Based on Deep Reinforcement Learning},
  journal = {Agronomy},
  volume  = {12},
  number  = {9},
  pages   = {2157},
  year    = {2022}
}

@inproceedings{R12,
  author    = {Y. Deng and X. Guo and Y. Wei and K. Lu and B. Fang and D. Guo and H. Liu and F. Sun},
  title     = {Deep Reinforcement Learning for Robotic Pushing and Picking in Cluttered Environment},
  booktitle = {Proc. IEEE/RSJ Int. Conf. Intelligent Robots and Systems (IROS)},
  pages     = {619--626},
  year      = {2019}
}

@inproceedings{R13,
  author    = {J. Matas and S. James and A. J. Davison},
  title     = {Sim-to-Real Reinforcement Learning for Deformable Object Manipulation},
  booktitle = {Proc. Conf. Robot Learning (CoRL)},
  pages     = {734--743},
  year      = {2018}
}

@inproceedings{R14,
  author    = {X. Lin and Y. Wang and J. Olkin and D. Held},
  title     = {SoftGym: Benchmarking Deep Reinforcement Learning for Deformable Object Manipulation},
  booktitle = {Proc. Conf. Robot Learning (CoRL)},
  pages     = {432--448},
  year      = {2021}
}

@inproceedings{R15,
  author    = {X. Zhang and S. Gupta},
  title     = {Push Past Green: Learning to Look Behind Plant Foliage by Moving It},
  booktitle = {Robotics: Science and Systems (RSS)},
  year      = {2024}
}

@inproceedings{R19,
  author    = {A. Wagenmaker and K. Huang and L. Ke and K. Jamieson and A. Gupta},
  title     = {Overcoming the Sim-to-Real Gap: Leveraging Simulation to Learn to Explore for Real-World RL},
  booktitle = {Advances in Neural Information Processing Systems (NeurIPS)},
  year      = {2024}
}

@article{R20,
  author  = {M. Mittal and C. Yu and Q. Yu and J. Liu and N. Rudin and D. Hoeller and J. L. Yuan and R. Singh and Y. Guo and A. Garg},
  title   = {Orbit: A Unified Simulation Framework for Interactive Robot Learning Environments},
  journal = {IEEE Robotics and Automation Letters},
  volume  = {8},
  number  = {6},
  pages   = {3740--3747},
  year    = {2023}
}

@article{R_LightStrawberry2026,
author = {M. Sun and S. Hu and C. Zhao and Y. Xiong},
title = {Light-Resilient Visual Regression of Strawberry Ripeness for Robotic Harvesting},
journal = {Computers and Electronics in Agriculture},
volume = {241},
pages = {111169},
year = {2026}
}

@article{R21,
  author  = {J. Schulman and F. Wolski and P. Dhariwal and A. Radford and O. Klimov},
  title   = {Proximal Policy Optimization Algorithms},
  journal = {arXiv preprint arXiv:1707.06347},
  year    = {2017}
}

@article{R23,
  author  = {G. Xu and R. Zheng and Y. Liang and X. Wang and Z. Yuan and T. Ji},
  title   = {DRM: Mastering Visual Reinforcement Learning Through Dormant Ratio Minimization},
  journal = {arXiv preprint arXiv:2310.19668},
  year    = {2023}
}

@inproceedings{R24,
  author    = {J. Luo and Z. Hu and C. Xu and Y. L. Tan and J. Berg and A. Sharma and S. Schaal and C. Finn and A. Gupta and S. Levine},
  title     = {SERL: A Software Suite for Sample-Efficient Robotic Reinforcement Learning},
  booktitle = {Proc. IEEE Int. Conf. Robotics and Automation (ICRA)},
  pages     = {16961--16969},
  year      = {2024}
}

@article{R25,
author = {N. Subedi and H.-J. Yang and D. K. Jha and S. Sarkar},
title = {Find the Fruit: Zero-Shot Sim2Real RL for Occlusion-Aware Plant Manipulation},
journal = {arXiv preprint arXiv:2505.16547},
year = {2026}
}

@inproceedings{R_HybridArm_IROS2024,
  author    = {Y. Chen and Z. Miao and Y. Ge and S. Lin and L. Chen and Y. Xiong},
  title     = {Design and Control of a Novel Six-Degree-of-Freedom Hybrid Robotic Arm},
  booktitle = {2024 IEEE/RSJ International Conference on Intelligent Robots and Systems (IROS)},
  address   = {Abu Dhabi, United Arab Emirates},
  year      = {2024},
  pages     = {3597--3604},
  doi       = {10.1109/IROS58592.2024.10802044}
}

\end{CJK*}
\end{document}